\documentclass{article}

\usepackage{arxiv}

\usepackage[utf8]{inputenc} % allow utf-8 input
\usepackage[T1]{fontenc}    % use 8-bit T1 fonts
\usepackage{hyperref}       % hyperlinks
\usepackage{url}            % simple URL typesetting
\usepackage{booktabs}       % professional-quality tables
\usepackage{amsfonts}       % blackboard math symbols
\usepackage{nicefrac}       % compact symbols for 1/2, etc.
\usepackage{microtype}      % microtypography
\usepackage{lipsum}		% Can be removed after putting your text content
\usepackage{graphicx}
\usepackage{natbib}
\usepackage{doi}
\usepackage{amsmath}
\usepackage{rotating}
\usepackage{booktabs}  % For better table formatting
\usepackage{geometry}  % To adjust the page margins if needed
\usepackage{multirow}  % For multi-row cells
\title{Can large language models replace humans in the systematic review process? Evaluating GPT-4’s efficacy in screening and extracting data from peer-reviewed and grey literature in multiple languages}

\author{ \href{https://orcid.org/0000-0002-4664-1204}{\includegraphics[scale=0.06]{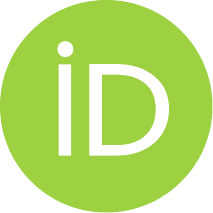}\hspace{1mm}Qusai Khraisha} \\
	Trinity Centre for Global Health\\
        School of Psychology\\
	Trinity College Dublin\\
	Ireland \\
	\texttt{khraishq@tcd.ie} \\
	\And
	{\hspace{1mm}Sophie Put} \\
	Department of Education\\
	University of York\\
	UK \\
 \And
	{\hspace{1mm}Johanna Kappenberg} \\
        School of Psychology\\
	Trinity College Dublin\\
	Ireland \\
 \And
	{\hspace{1mm}Azza Warraitch} \\
        Trinity Centre for Global Health\\
        School of Psychology\\
	Trinity College Dublin\\
	Ireland \\
 \And
	{\hspace{1mm}Kristin Hadfield} \\
        Trinity Centre for Global Health\\
        School of Psychology\\
	Trinity College Dublin\\
	Ireland \\
}

\renewcommand{\shorttitle}{\text Can large language models replace humans in the systematic review process?}

\hypersetup{
pdftitle={GPT paper},
pdfsubject={q-bio.NC, q-bio.QM},
pdfauthor={Qusai Khraisha},
    pdfkeywords={Systematic reviews, Large language models, LLMs, GPT, Artificial intelligence, AI, Natural Language Processing, NLP, Machine learning}
}

\begin{document}
\maketitle

\begin{abstract}
Systematic reviews are vital for guiding practice, research, and policy, yet they are often slow and labour-intensive. Large language models (LLMs) could offer a way to speed up and automate systematic reviews, but their performance in such tasks has not been comprehensively evaluated against humans, and no study has tested GPT-4, the biggest LLM so far. This pre-registered study evaluates GPT-4’s capability in title/abstract screening, full-text review, and data extraction across various literature types and languages using a ‘human-out-of-the-loop’ approach. Although GPT-4 had accuracy on par with human performance in most tasks, results were skewed by chance agreement and dataset imbalance. After adjusting for these, there was a moderate level of performance for data extraction, and – barring studies that used highly reliable prompts – screening performance levelled at none to moderate for different stages and languages. When screening full-text literature using highly reliable prompts, GPT-4’s performance was ‘almost perfect.’ Penalising GPT-4 for missing key studies using highly reliable prompts improved its performance even more. Our findings indicate that, currently, substantial caution should be used if LLMs are being used to conduct systematic reviews, but suggest that, for certain systematic review tasks delivered under reliable prompts, LLMs can rival human performance. 
\end{abstract}

% keywords can be removed

\section{Introduction}
Systematic reviews play a crucial role in advancing practice, research, and policy (Aromataris et al., 2015). However, the current approach to systematic reviews is laborious and can be slow to the point that the resulting synthesis of knowledge may no longer be up to date when it is completed (Borah et al., 2017; Michelson and Reuter, 2019). The explosion of scientific literature, coupled with the complexity and specificity of many research questions, further adds to these challenges (Fiorini et al., 2018). Artificial intelligence (AI) has emerged as a potential solution to these challenges, with recent studies and evaluations suggesting its capability to enhance the quality and efficiency of systematic reviews (Blaizot et al., 2022; Dijk et al., 2023; Kebede et al., 2023; Mahuli et al., 2023; Moreno-Garcia et al., 2023; Nugroho et al., 2023; Santos et al., 2023).

Some examples of AI tools that have been used in systematic reviews include Rayyan and Abstracker, which help with screening titles and abstracts (Giummarra et al., 2020; Rogers et al., 2020), trialStreamer, which helps with data extraction from full-text articles (Marshall et al., 2020), and RobotReviewer, which helps with assessing study quality and bias (Goldkuhle et al., 2018). The major shortcoming of these tools is that their performance significantly deteriorates without looping in a human in the decision-making process, as shown by Blaizot and colleagues' (2022) meta-analysis. One possible reason for these limitations is that they split text into fixed segments, which has been argued to hinder their ability to understand context, especially in longer texts (Guo et al., 2023). Newer large language models (LLMs) based on the transformer technology (Vaswani et al., 2017), such as Generative Pre-Trained Transformer (GPT) and Bidirectional Encoder Representations from Transformers (BERT), may overcome this problem since they capture more contextual information. This was demonstrated in a recent study, which highlighted GPT’s superior performance on systematic review tasks compared to older AI methods (Syriani et al., 2023). 

The question of whether AI tools can match or surpass human performance in conducting systematic reviews carries profound implications for the future of scientific research. It holds the potential to radically transform knowledge synthesis, turning systematic reviews from static literature summaries into dynamic, continually updated resources – potentially altering the very way we approach science. Given these significant implications, it is crucial to acknowledge the current uncertain state of AI in this domain, especially regarding the most substantial LLM model, GPT-4. This model has been reported to significantly surpass all other LLMs, including previous versions of GPT, in various natural language processing tasks across both English and other languages (OpenAI, 2023). Yet, as of now, nothing is documented about GPT-4’s performance in conducting systematic reviews.

Research on using other LLMs in systematic reviews (mostly earlier versions of GPT) is not as comprehensive or systematic as it could be, with much of the work containing contaminated datasets and inadequate metrics. No study, for instance, has tested grey literature and non-English literature, which can constitute a large proportion of the evidence base for some topics (Lawrence et al., 2014). Most studies focused on narrow aspects of systematic reviews, such as Boolean queries (S. Wang et al., 2023) or only evaluating performance on titles and abstracts screening (Alshami et al., 2023; Guo et al., 2023; Syriani et al., 2023). Some have methodological shortcomings, such as Mahuli et al., (2023), who did not provide an objective evaluation of GPT’s performance, or Alshami et al., (2023), who did not test GPT’s autonomous performance, instead relying on a ‘human-in-the-loop’ approach. A few may have included contaminated data, such as Guo et al., (2023) and Syriani et al., (2023), who used datasets for systematic reviews that published their results before or in 2021, when GPT was trained, which may bias the results in favour of GPT. Other studies, such as Guo et al., (2023), did not consider imbalance nor incorporate chance agreement in their interpretation of the results, thereby potentially inflating GPT’s accuracy. Syriani et al., (2023) addressed this issue but focused on investigating GPT’s performance against other AI tools, not human reviewers. Our pre-registered study is the first to evaluate GPT-4’s autonomous performance across several systematic review processes, including title/abstract screening, full-text screening, and data extraction. It is also the first to test an LLM model in reviewing grey literature and literature in other languages.

\section{Methods}
\label{sec:headings}
This study assessed GPT-4’s performance in screening and extracting data from documents for an ongoing systematic review on parenting in protracted refugee situations. We used the ChatGPT interface to access the GPT-4 model between May and September 2023. We tested documents that were reviewed using four inclusion/exclusion criteria: containing empirical data, parenting behaviour, refugee status, and protracted refugee situation. Links to our GPT-4 prompts and outputs, as well as the R code used for analysis, are on the Open Science Foundation (OSF) page (link). We registered this study protocol on the OSF, while the details of the review can be found on both OSF (link) and PROSPERO (Anonymised). We screened 300 titles/abstracts and 150 full-texts, as well as extracted data from 30 documents (see Figures 1 and 2). Sample size was largely based on studies reviewed by at least two humans for screening, which was the case for English language documents at all stages and those written in other languages in the title/abstract stage. However, due to time and resource constraints, we only used one human reviewer for non-English studies at the full-text level, and for data extraction from all studies. While we ensured a mix of decisions in terms of a random selection of relevant and irrelevant studies written in English to gain deeper insight into inclusion performance, this was generally not possible for non-English studies due to the large number of excluded studies. 

\begin{figure}[h]
    \centering
    \includegraphics[width=1\textwidth]{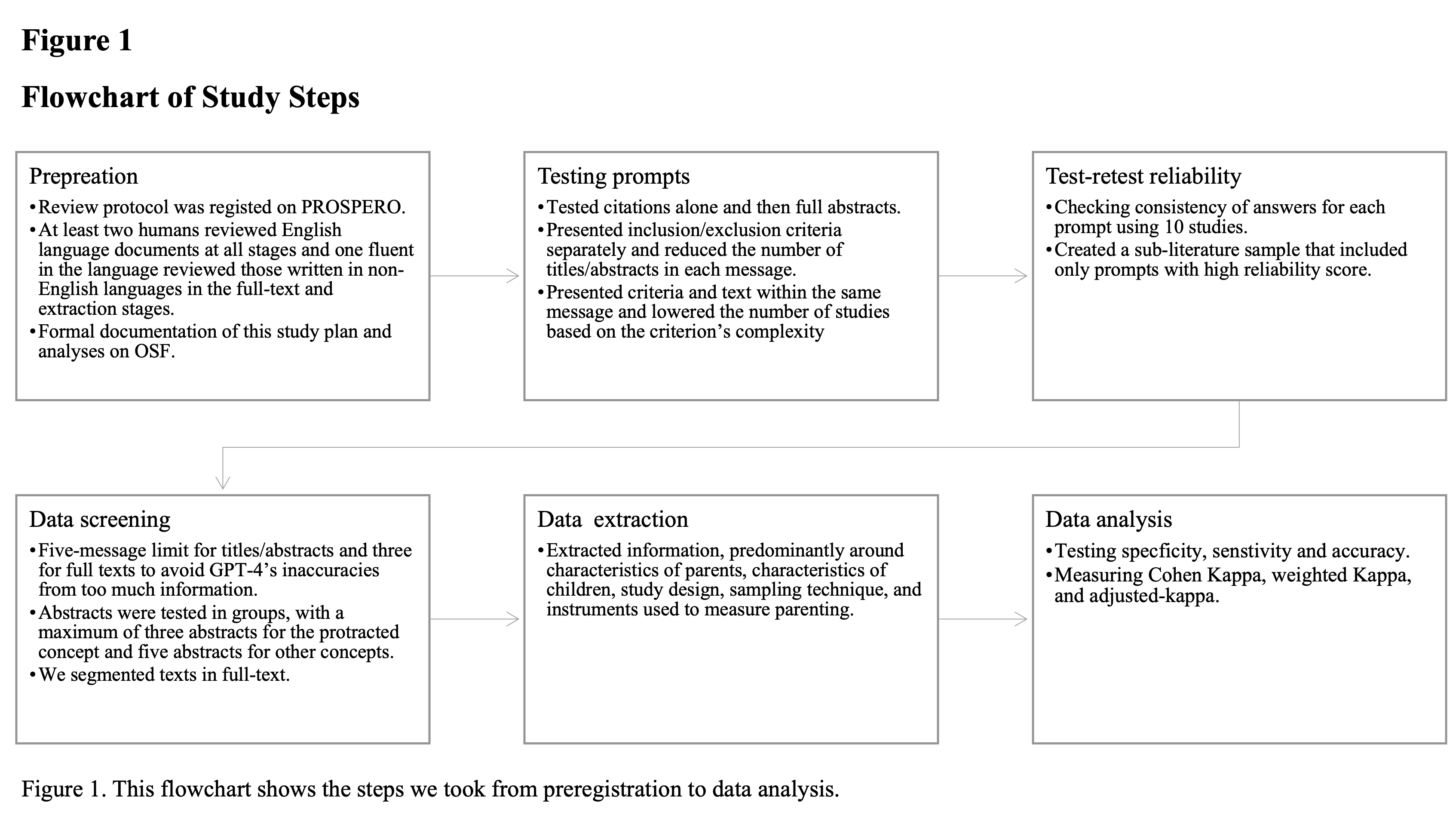}
    \label{fig:Picture_1}
\end{figure}

\begin{figure}[h]
    \centering
    \includegraphics[width=1\textwidth]{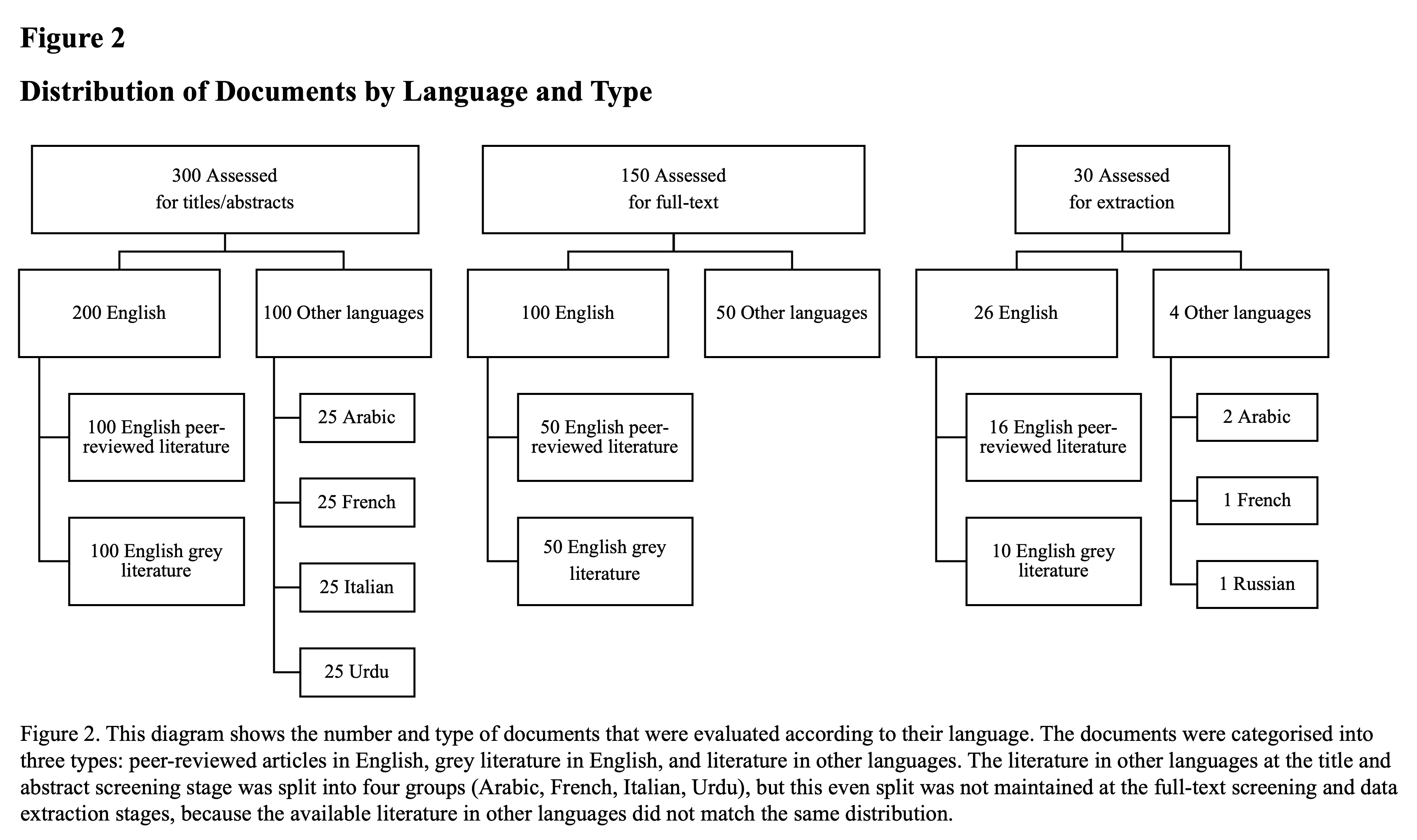}
    \label{fig:Picture_2}
\end{figure}

\subsection{Prompt engineering approach }
We experimented with GPT-4 prompt formats for title/abstract screening. Initially, we tested citations alone, but GPT-4 appeared to focus mainly on titles, and its accuracy decreased with an increased volume of citations. To address this, we presented complete abstracts in our prompts. Our next challenge was finding that GPT-4 struggled with complex queries and large data volumes, prompting us to present inclusion/exclusion criteria separately and reduce the number of titles/abstracts in each message. This approach seemed to reduce hallucinations, a frequently observed phenomenon where an LLM confidently produces an inaccurate output (Beutel et al., 2023), but reduced consistency upon retesting. Presenting criteria and text within the same message and lowering the number of studies based on the criterion’s complexity improved consistency without increasing hallucinations. 

We assessed test-retest reliability using 10 studies on each of our four criteria. Each criterion corresponds to a prompt. We tested the four prompts five times and generated five decisions per criterion per study. We then recorded each time GPT-4 output was inconsistent from the original answer. These scores were used to assess the impact of prompt reliability on accuracy scores.

\subsection{Screening and extraction}
Each abstract batch (a maximum of three abstracts for the protracted concept – given that it contained a longer prompt – and five abstracts for other concepts) underwent four chat tests based on inclusion/exclusion criteria. We tested multiple abstracts within the same query to increase efficiency in screening. We proceeded to the next criterion in another chat if GPT-4 responded ‘yes/maybe’. A ‘no’ meant exclusion from further review. If an abstract was excluded for not meeting a criterion, we removed it and introduced a new one for subsequent tests, meaning that the number of abstracts remained the same every time a criterion was tested. We set a five-message limit for titles/abstracts and three for full texts to avoid GPT-4’s inaccuracies arising from too much information. Lengthy abstracts in grey literature were divided for separate evaluations until GPT-4 finalised a decision. During grey literature screening, we modified the query to fit source formats, switching “titles/abstracts” to “websites/reports” and “texts.” 

For full text, we adjusted our queries, using “Include” instead of “Yes/maybe” and “Exclude” for “No”. Due to character limits, we segmented texts. To reduce contamination of responses with each other, every time a text snippet was sent to GPT-4, it was accompanied by the specific inclusion/exclusion criterion it was being tested against together within the same prompt. If GPT-4 said “Include” for a segment, we viewed that criterion as met and began a new chat with the next criterion from the first segment. An “Exclude” led us to present subsequent segments. If GPT-4 never indicated “Include” by the last segment, the study was marked as excluded, and no further criteria were tested. If humans had already excluded a study, GPT-4 began screening with the humans’ exclusion reason. If GPT-4 shared the decision made by humans, no further tests were needed. Otherwise, all remaining criteria were checked until we received an exclusion decision. As a result, 72\% of the English peer-reviewed literature needed testing on the ‘parenting’ and ‘protracted refugee situations’ prompt, compared to 44\% and 18\% for grey and non-English literature, respectively.

Extracted information mainly revolved around characteristics of parents (e.g., number of parents, gender, and education distribution), characteristics of children (e.g., number of children, gender, and education distribution), study design (by duration, data collection method and group allocation), sampling technique, and instruments used to measure parenting (e.g., instrument type and name, target respondent and the main focus of the instrument). Only text from the methods and results sections of the chosen papers were inputted into GPT-4. Similar to the above, we segmented the text into pieces. Only the first response to a prompt was deemed final, except if the response was incomplete, then the subsequent responses were also recorded as part of the answer. 

\subsection{Analysis metrics}
We applied four metrics: True Positives (TP), True Negatives (TN), False Positives (FP), and False Negatives (FN). TP and TN are the cases where GPT-4 agreed with human reviewers, meaning it made correct decisions. FP and FN are the cases where GPT-4 disagreed with human reviewers, meaning it made wrong decisions. Ratios derived from these metrics provide insight into the imbalance of the dataset by dividing true cases against false cases. A meta-analysis indicated that systematic reviews’ datasets have between 40\% to less than 1\% of relevant studies, with an average of 3\%. This suggests that imbalance scores of 3\% are typical, while a score of 40\% or less than 1\% is ‘somewhat typical,’ and a score of above 40\% is ‘atypical’ (Sampson et al., 2011).

\begin{equation}
\text{Imbalance} = \frac{TP+FP}{TN+FN}
\end{equation}

For GPT-4 performance, we calculated sensitivity, specificity and accuracy, which are commonly used metrics (e.g., Frömke et al., 2022; Patel et al., 2021). Sensitivity (also known as recall) shows how well GPT-4 identified positive cases by taking (TP)  and dividing them by the sum of TP and FN. Specificity indicates how well GPT-4 identified negative cases by taking TN and dividing them by the sum of TN and FP. Accuracy, which evaluates the overall correctness of GPT-4, was calculated by adding TP and TN and dividing by the total number of cases. There is no consensus on how to interpret these scores, as they largely depend on the context (e.g., Shreffler and Huecker, 2023). Previous studies have reported that human error rates in systematic review screening range from 5\% to 20\%, implying that a score of 100\% is ‘superior’ to humans, while an accuracy score of 80\% to 95\% for GPT-4 could be regarded as ‘on par’ (Wang et al., 2020). In the worst possible documented prediction, human error rates reached up to 40\% (Wang et al., 2020). This suggests that accuracy scores between 60\% and 80\% could be regarded as ‘near-par,’ and scores below 60\% could be regarded as ‘subpar’ to humans.

\begin{align}
\text{Sensitivity} & = \frac{TP}{TP+FN} \\
\text{Specificity} & = \frac{TN}{TN+FP} \\
\text{Accuracy} & = \frac{TP+TN}{TP+TN+FP+FN}
\end{align}

Cohen’s Kappa (\( \kappa \)) was calculated to compare the actual and expected agreement between GPT-4 and human reviewers (Cohen, 1960). This is important for systematic reviews, where inclusion is rare, and exclusion is common, which can make humans and GPT-4 seem more agreeable than they are. The actual agreement (Po) is the proportion of cases humans and GPT-4 gave the same rating, positive or negative. The expected agreement (Pe) is the probability that humans and GPT-4 gave the same rating by chance, based on their rating frequencies. We subtracted the expected agreement from the actual agreement and divided it by the maximum possible agreement (1 minus the expected agreement). This gives a score between -1 and 1. We followed what McHugh (2012) suggested for the classification of agreement scores, which are more stringent on interpreting values than those suggested by Cohen (1960): values .0 – .20 as no agreement, .21 – .40 as minimal, .41 – .59 as weak, .60 –.79 as moderate, and 0.80 – .90 as almost perfect agreement. 

Cohen’s kappa has been shown to produce a false agreement rate in imbalanced datasets, so we used PABAK (prevalence-adjusted bias-adjusted kappa) to account kappa for the effects of prevalence and bias in the data set (Byrt et al., 1993). PABAK corrects for this by accounting for the distribution of the categories in the denominator and by adjusting the counts of agreements and disagreements in the formula. We also used weighted kappa to better capture the severity of false rejections by GPT-4, which are the most serious errors it can make, because this would exclude a study which meets the inclusion criteria. Weighted kappa (\(\omega\kappa\)) assigns higher weights to greater disagreements, with 0 for complete agreement. There is no standard test that can combine weights with PABAK, so we could not account for the effects of data imbalance in the weighted Cohen’s kappa scores. Sampson et al. (2011) found that the median search precision for systematic reviews is around 3\%; that is, there are about 30 times more excluded studies than included ones. This suggests a weight of 30 for false rejections in the calculation of weighted kappa. 

\begin{align}
\kappa & = \frac{1 - \text{Pe}}{\text{Po} - \text{Pe}} \\
\omega\kappa & = 1 - \frac{1 - \text{Po}\omega}{1 - \text{Pe}\omega} \\
\text{PABAK} & = \frac{1 - 0.5}{2\text{Po} - 1}
\end{align}

\section{Results}
\label{sec:headings}
In this study, we evaluated GPT-4’s performance in screening and extracting peer-reviewed, grey (non-peer reviewed), and non-English literature. We first report on the reliability of answers when using the same prompt. This means that GPT-4 gave the same answer every time, without any variation. For instance, if GPT-4 gave the same answer 10 times in a row for the same text and prompt, it scored 100\%, but if it gave the same initial answer only 7 times out of 10, it scored 70\%. GPT-4 performed best when assessing empirical data and refugees (100\% reliability) and struggled with the concepts of parenting behaviour (50\% reliability) and protracted refugee situations (70\% reliability). With these findings in mind, we created a sub-literature sample that included only prompts relating to refugee status and empirical data called the ‘high-reliability prompt group,’ given that these prompts had the highest reliability scores. Ultimately, this sub-sample included 23 studies: a third were English peer-reviewed studies, a third were English grey literature studies, and another third were non-English studies. 

We subsequently looked at the balance of data, which is the ratio of relevant to irrelevant studies, for each literature type, language, and stage (for the extraction stage, this means the presence or absence of data). Peer-reviewed studies in English were fully balanced, as intended by our design, except in the extraction stage (.03; that is, 1 included for every 30 excluded). Unlike non-English studies (.05), the grey literature was fully balanced at the title/abstract stage, but then both grey literature and non-English studies were skewed towards irrelevance in the full-text screening (.11 and .09, respectively) and extraction stages (.24 and .20, respectively). It is important to note that while balance aids in understanding all aspects of performance, it does not reflect the inherent imbalances in real-world datasets of systematic review. Based on Sampson et al., (2011) finding that there are typically about 30 times more excluded studies than included studies, our most imbalanced datasets were also the most consistent with other systematic reviews.

\begin{sidewaystable}
\centering
\caption{Performance Evaluation of GPT-4 versus Human Reviewers in Screening and Extraction}
\begin{tabular}{@{}lccccccc@{}}
\toprule
 & Balance & Sensitivity & Specificity & Accuracy & Cohen Kappa$\ast$ & Weighted Kappa & Adjusted Kappa$\ast\ast$ \\
 & & & & & & & \\
\multicolumn{8}{@{}l}{\textbf{Title and abstract screening}} \\
\midrule
English peer-reviewed & 1 & .42 & .92 & .67 & .34 & .23 & .34 \\
English grey & 1 & .48 & .84 & .66 & .32 & .24 & .32 \\
Other languages & .05 & .50 & .89 & .88 & .21 & .40 & .75 \\
\multicolumn{8}{@{}l}{\textbf{Full text screening}} \\
\midrule
English peer-reviewed & .92 & .38 & .69 & .54 & .07 & .05 & .08 \\
English grey & .11 & .60 & .80 & .78 & .24 & .44 & .55 \\
Other languages & .09 & 1 & .95 & .96 & -.10 & -.11 & .64 \\
\multicolumn{8}{@{}l}{\textbf{High-reliability prompt group}} \\
\midrule
High-reliability prompt group & .05 & .36 & .94 & .85 & .65 & .97 & .91 \\
\multicolumn{8}{@{}l}{\textbf{Data extraction}} \\
\midrule
English peer-reviewed & .03 & .75 & .84 & .82 & .54 & .63 & .63 \\
English grey & .24 & .65 & .85 & .81 & .45 & .53 & .62 \\
Other languages & .20 & .36 & .94 & .85 & .35 & .29 & .69 \\
\bottomrule
\end{tabular}

\textbf{*}  The inter-rater reliability between human reviewers for the full-text data was a Cohen Kappa coefficient of .77. However, the review is not yet completed, so the final value may vary slightly. The inter-rater reliability between human reviewers was calculated using data mostly from the peer-reviewed literature, but it also includes a small portion of grey literature and non-English studies.

\textbf{**} The human reviewers achieved an adjusted Cohen Kappa of .89 for the same literature sample Cohen Kappa was calculated for above.

Note. We used PABAK (Prevalence-Adjusted Bias-Adjusted Kappa) to adjust Kappa for the effects of prevalence and bias in the data set (Byrt et al., 1993) and a weight of 30 for false rejections in the calculation of weighted Kappa based on previous studies which have shown that the median search precision for systematic reviews is around 3\% (Sampson et al., 2011).
\end{sidewaystable}

Across all stages and categories, the specificity was on par with human performance (>.80, except for English peer-reviewed full-text screening), indicating a robust ability of GPT-4 to correctly identify irrelevant studies (Table 1). This was especially true in literature containing non-English studies (>.90). Sensitivity, indicating how effectively GPT-4 identified relevant studies, was highest in the extraction stage for both peer-reviewed (English: .75) and grey literature (.65), although note that for non-English studies the sensitivity was only .36 for data extraction. For non-English studies, perfect sensitivity was achieved during the full-text stage. Accuracy was somewhat higher in the data extraction stage than in the title/abstract screening phase, ranging between near-par and on par with human performance, except for the English peer-reviewed literature.

The balanced dataset of English peer-reviewed literature had lower accuracy (title/abstract: .67, full-text: .69) than the more unbalanced non-English literature dataset (title/abstract: .88, full-text: .96). In extraction, which was the only time the dataset of English peer-reviewed literature was imbalanced, accuracy was the highest (.84). This suggests that GPT-4’s high accuracy might be due to chance. Supporting this notion, the associated adjusted kappa scores were low, ranging from ‘none’ to ‘moderate’ as categorised by McHugh (2012). An outlier in these scores was the ‘almost perfect’ agreement seen in the highly-reliable prompt group, which exclusively featured responses from highly reliable prompts (.91). When we weighted kappa to emphasise false rejections, in a way penalising GPT-4 for missing key studies, scores for the highly reliable prompt group improved even more (.97).  

\section{Discussion}
\label{sec:headings}
 We found mixed results on the efficacy of GPT-4 as compared to human reviewers across various systematic review tasks, languages, and literature types. GPT-4’s accuracy was influenced by chance agreement and dataset imbalance, and when these factors were considered, GPT-4 often substantially underperformed humans. Yet, under specific conditions – namely, when given entirely highly reliable prompts in full-text screening – GPT-4 demonstrated an ‘almost perfect’ performance on par with humans. While our findings indicate the need for caution in assuming uniform proficiency across tasks, they also suggest that, under certain conditions, LLMs have the potential to revolutionise how we synthesise knowledge. 

 Our results for sensitivity, and specificity are consistent with previous studies. For instance, Guo et al. (2023) and Alshami et al. (2023) found that specificity was the strongest metric for title/abstract screening, as we did. They achieved a specificity score close to ours (90\% and 93\%, respectively, vs our 92\%). Their sensitivity score was slightly higher than ours (76\% and 84\%, respectively, vs our 67\%; although note that Alshami et al., 2023 used a human-in-the-loop method). Our accuracy score for the peer-reviewed literature (67\%) was lower than in previous work, possibly because our study was the only one that artificially balanced its dataset. Unlike the others (Alshami et al., 2023; Guo et al., 2023), we took this step because skewed datasets can make accuracy metrics unreliable, as later indicated by our low adjusted kappa. Such a limitation will not be easily detected by anecdotal tests (Mahuli et al., 2023; Qureshi et al., 2023) and may mislead general users who may assume that GPT is performing well when it is not. A possible reason for this misconception is that GPT typically generates text that resembles human writing in terms of quality, style, and content (Jakesch et al., 2023), thus misleading users to think that GPT has human-level abilities based on its human-like outputs.
 
 This study was the first to report on the novel application of an LLM in conducting full-text screening and extraction. Automation techniques have been hailed as a way of increasing reproducibility (Ivimey-Cook et al., 2023). However, previous endeavours employing AI tools for these tasks pinpointed several challenges. These included the need for continuous human intervention in full-text screening (Beller et al., 2018), the necessity for extensive pre-screening of training datasets (Halamoda-Kenzaoui et al., 2022), and the incapability to either review full-texts (Clark et al., 2020; Nye et al., 2018) or to extract data deviating from a predetermined structure (Summerscales et al., 2011; Wallace et al., 2016). Our findings indicate that GPT-4 has limited potential in full-text screening and data extraction, with moderate performance in non-English and grey literature, and a very poor ability with English peer-reviewed texts. Its training on publicly available data, which might lean more towards grey literature and non-English sources, could explain this difference. However, we should note that the English peer-reviewed literature data had a very unusual balance of studies, unlike the other databases, which are closer in their compositions to other systematic reviews, which suggests caution against assuming generalisability. Lastly, GPT-4 performance was strongly influenced by prompt reliability, which could itself be affected by word count and prompt complexity. Longer prompts, like those for parenting behaviour and protracted refugee situations (around 400 words and 1600 words, respectively; the other two prompts were less than 300 words), may have lost important context due to their size. We also observe that the ‘parenting behaviour’ prompt might have been most challenging for GPT-4 because the prompt was more open than specific (to capture non-traditional ways of parenting), unlike all other ones, which contained exact definitions.
 
There are three main strengths and weaknesses to this study. First, despite our comprehensive approach covering various literature types and our efforts to achieve balance, the challenges in balancing non-English texts and our very attempts at balancing could have both introduced biases. Second, we used various metrics to measure GPT-4’s accuracy, consistency, and agreement with human reviewers, but our relatively small sample size of evaluated papers might limit the generalisability of GPT-4’s findings for other systematic reviews. Third, we registered and documented our research protocol and analysis plan to ensure its validity and replicability but might have benefited from also creating a detailed plan for prompt engineering in advance.

\section{Conclusion}
\label{sec:headings}
Over a hundred years ago, audiences worldwide were captivated by a horse named Hans, who, with a confident tap of his hoof, appeared to solve mathematical problems. They believed Hans possessed an extraordinary cognitive ability, almost human-like in nature. Yet, beneath this illusion lay a simpler truth: Hans was reading his handler. He picked up on the faintest of cues – a twitch of a muscle, a barely noticeable nod, or even an unconscious sigh of anticipation. This phenomenon mirrors the workings of LLMs like GPT-4. While not deciphering human cues per se, they are heavily influenced by the prompts they receive, much like Hans needed clear guidance from his handler. But there is a key distinction: whereas Hans required human guidance for his output, GPT-4 needs clear human input and generates outputs autonomously. Our study underscores this, showing that when given a reliable prompt, GPT-4’s screening performance rises to be almost perfect. In harnessing this potential, LLMs might pave the way for a transformative era in systematic reviews.

%Bibliography
\nocite{*}

\bibliographystyle{apalike}

\bibliography{Khraisha_et_al}  % Uncomment this line and comment out the ``thebibliography'' section below to use the external .bib file (using bibtex).

\end{document}